\documentclass[sigconf,10pt,natbib=true]{acmart}
\usepackage{bigfoot}
\DeclareNewFootnote[para]{default}
\usepackage{tikz}
\usetikzlibrary{arrows, decorations.markings,shapes,arrows,fit, positioning}
\usetikzlibrary{shadows}
\usepackage{tablefootnote}
\usepackage{multirow}
\usepackage{booktabs}
\usepackage[normalem]{ulem}
\useunder{\uline}{\ul}{}
\usepackage{float}
\usepackage{placeins}
\usepackage[tablename=Table]{caption} 
\usepackage{enumitem}
\usepackage{varwidth}
\usepackage{tasks}
\def\threedigits#1{%
  \number#1}
\usepackage[justification=centering]{caption}
\renewcommand\footnotetextcopyrightpermission[1]{}
\AtBeginDocument{%
  }

\setcopyright{none}
\begin{document}


\title{Characterizing Latent Perspectives of Media Houses Towards Public Figures}
\author{Sharath Srivatsa}
\affiliation{
  \institution{International Institute of Information Technology}
  \city{Bangalore}
  \country{India}
}
\email{sharath.srivatsa@iiitb.org}

\author{Srinath Srinivasa}
\affiliation{%
  \institution{International Institute of Information Technology}
  \city{Bangalore}
  \country{India}
}
\email{sri@iiitb.ac.in}

\begin{abstract}
Media houses reporting on public figures, often come with their own biases stemming from their respective worldviews. A characterization of these underlying patterns helps us in better understanding and interpreting news stories. For this, we need diverse or subjective summarizations, which may not be amenable for classifying into  predefined class labels. This work proposes a zero-shot approach for non-extractive or generative characterizations of person entities from a corpus using GPT-2. We use well-articulated articles from several well-known news media houses as a corpus to build a sound argument for this approach. First, we fine-tune a GPT-2 pre-trained language model with a corpus where specific person entities are characterized. Second, we further fine-tune this with demonstrations of person entity characterizations, created from a corpus of programmatically constructed characterizations. This twice fine-tuned model is primed with manual prompts consisting of entity names that were not previously encountered in the second fine-tuning, to generate a simple sentence about the entity. The results were encouraging, when compared against actual characterizations from the corpus.
\end{abstract}

\maketitle
\pagestyle{plain}
\section{Introduction}
Media houses have their own worldview with which they interpret happenings around the world, that may show up as biases in their characterizations of entities like persons, organizations or countries. Such biases are often implicit and benign. However, in order to get better clarity and understanding of news, it is important to explicate and understand how specific media houses characterize specific entities. 

Automated approaches for entity characterizations have gained significant interest in recent years~\cite{wei2019esa, liu2020deeplens,liu2021entity}. Most of the current approaches are extractive in nature, that look for specific features like frequency, diversity, informativeness, etc. in the descriptions of entities to extract sentences characterizing them. 

Given the vast numbers of entities and issues that media houses report on, it is impractical to create a pre-determined set of classes onto which, characterizations can be classified. We define entity characterization as a terse or one statement description of the individual quality of a person or a thing. In contrast to summarization, entity characterization need cover all pertinent characteristics of the entity in a short summary. Entity characterization is subjective-- that reveals the biases of the inquirer, whereas summarization is meant to be objective, and verifiable against the actual characteristics of the object of inquiry. 

In this work, we propose an approach for \textit{non-extractive} or \textit{generative} characterizations for person entities. These are in the form of one-sentence descriptions that are obtained from suitably fine-tuning pre-trained masked language models. 

Pre-trained masked language models are known to perform well on diverse NLP tasks in a zero-shot setting~\cite{GPT2}. A number of recent approaches~\cite{schick2021s,schick2020automatically,hambardzumyan-etal-2021-warp,gao-etal-2021-making,petroni2019language} show that domain-adapted language models have substantial knowledge of the domain, and with pattern demonstrations to solve a NLP task, the model performs well in the intended task. Our approach is along the same lines to perform Generative Entity Characterization by fine-tuning with demonstrations.

Our approach is to fine-tune the GPT-2 pre-trained language model twice, and use this model to generate characterizations. The first fine-tuning is for domain adaptation with a corpus of person entity mentions disambiguated for co-references. For the second fine-tuning we perform \textit{entity characterization demonstrations}, based on  sentences characterizing the entity in question. These sentences are programmatically constructed from the corpus by extracting clauses and their parts. Subject, Verb, Object, and Adverbials are the common parts of clauses that are extracted. We construct a demonstration pattern to convert parts of clauses into semantically coherent simple sentences describing the entity. The pattern is to suffix the subject with \textit{``is described as''}, convert lemmatized verb into a gerund, and append other parts grammatically. With this pattern, a corpus of simple sentences about entities is constructed. Demonstration sentences for ten entities are then separated from this corpus to be used for testing, and are not included the demonstrations training. 

The twice fine-tuned model is then prompted with test entities suffixed with four different manual prompts, and the generated texts were inspected for characterizations of entities. Since the test entity sentences were not used in demonstrations, we attribute the generated text to zero-shot generations.

\section{Related Work}
In the recent past, text classification with language models and pattern training has shown promising results on key datasets. \citeauthor{schick2021exploiting}~\cite{schick2021exploiting}, show that language models understand text classification task by converting input to \textit{cloze question} patterns and training. Substantial soft-labeled inputs are generated with semi-supervised and ensemble language models to train the final classifier. Good results are observed with zero to few numbers of the initial dataset.


GPT-3 with hundreds of billions of parameters shows remarkable few-shot performance on SuperGLUE. \citeauthor{schick2021s} \cite{schick2021s} show that an equivalent few-shot performance can be achieved by training small language model ALBERT with \textit{cloze question} patterns. A set of SuperGLUE tasks require \textit{cloze question} patterns with multiple masks, a strategy to handle this is shown.

In the text classification task, mapping predicted tokens to predefined labels is challenging and requires domain expertise even though training with patterns optimizes text classification. \citeauthor{schick2020automatically} \cite{schick2020automatically}, show an approach to automatically map the predicted tokens to labels. Training language models with patterns have shown adequate performance in the text classification task. In this work, we propose a similar approach with manual prompts patterns to generate non-extractive information about person entities from a corpus.

Choosing prompts and equivalent words of classification labels manually or algorithmically are challenging since there are significant variations. \citeauthor{hambardzumyan-etal-2021-warp} \cite{hambardzumyan-etal-2021-warp} show an approach to finding these as embeddings in a continuous embedding space of word embeddings. Trainable embeddings are added around the input to make the masked language model predict the masked token and evaluated on natural language understanding tasks of GLUE Benchmark.

With natural language prompts and a few demonstrations on GPT-3, awe-inspiring performance on language understanding tasks is observed. However, since GPT-3 has 175B parameters, it is challenging to use in real-world applications. \citeauthor{gao-etal-2021-making} \cite{gao-etal-2021-making} show prompt-based fine-tuning with demonstrations on moderately small language models BERT and RoBERTa.  In this work, we have fine-tuned with person entity characterizing sentences as demonstrations.

Mining commonsense knowledge is an important natural language processing task. Language models are known to have this, \citeauthor{davison-etal-2019-commonsense} \cite{davison-etal-2019-commonsense} show an approach to mine commonsense knowledge from Pre-trained Language Modes. A uni-directional model generates sentences with a specific template for each type of relation in information triples. This generated sentence is validated by masking and predicting the tokens using a bi-directional language model.

Apart from linguistic knowledge, language models might also contain relational knowledge in the training data. \citeauthor{petroni2019language} \cite{petroni2019language}  analyze relational knowledge in state-of-the-art pre-trained language models with LAMA (LAnguage Model Analysis) probe a corpus of facts in subject-relation-object triples or question-answer pairs forms derived from diverse factual and commonsense knowledge sources.
\citeauthor{kassner2020negated} \cite{kassner2020negated}, show that the ability of pre-trained language models to learn factual knowledge is not as good as humans learn by probing for facts with Negated LAMA and Misprimed LAMA. Ideally, these probe variants should result in contradictions, whereas it was not so, suggesting that factual knowledge extraction is based on pattern matching rather than inference.
\citeauthor{jiang2020x} \cite{jiang2020x} study factual knowledge in multilingual language models with manually created probes in 23 languages similar to LAMA.

\citeauthor{DBLP:conf/acl/KumarT21} \cite{DBLP:conf/acl/KumarT21}, show that the order of training examples significantly reduces the samples required for few-shot learning on Sentiment Classification, NLI, and Fact Retrieval tasks.

\citeauthor{nishida2020unsupervised} \cite{nishida2020unsupervised} shows an approach where the pre-trained BERT is adapted to the target domain and next fine-tuned with RC task on a source domain. Finally, this model performs RC tasks in the target domain. The key idea is that the model is trained for a task on one domain and used to perform the task on another model.

Domain adaptation is crucial to solving any task related to that domain.\citeauthor{gururangan2020don} \cite{gururangan2020don} show that even pre-trained language models of hundreds of millions of parameters are ineffective to encode the nuances of a given textual domain. Therefore, it is necessary to specialize the model with the domain or task-relevant corpus before solving any task in a textual domain.

The state-of-the-art of using pre-trained language models to solve an NLP task show domain adaptation and fine-tuning with demonstrations of patterns as the most plausible approach to a reasonable extent. In this work, we propose an approach to characterize entities along similar lines.

\section{GPT-2 Domain Adaptation}
A GPT-2 Pre-trained Language Model (PLM), with 345M parameters, was fine-tuned with steps from GitHub.\footnote{GPT-2 Fine-tuning: \url{https://github.com/openai/gpt-2/}} PLM was fine-tuned individually on four popular news media corpora. Due to limitations in the available compute instance, 345M PLM, medium model was fine-tuned and this model proved sufficient to get convincing results. Domain adaptation or fine-tuning PLM on domain corpora is a prerequisite before task-specific training. The domain-adapted PLM was further fine-tuned with programmatically constructed demonstration sentences.

\subsection{Textual Media Source}
The GDELT Project\footnote+{GDELT Project: url{https://www.gdeltproject.org/}} records the world's broadcast, print, and web news from nearly every corner of every country in over 100 languages. From GDELT database, textual news media article URLs of four popular media houses, between years 2015 to 2021, were extracted and texts of articles were scraped for domain adaptation. Table \ref{tab:scrapedArticlesDetails} shows the details of each media house corpus.
\begin{table}[!htbp]
\centering
\caption{Scraped Media House Articles between \\ 2015 and 2021}
\label{tab:scrapedArticlesDetails}
\begin{tabular}{ccc} 
\toprule
\textbf{Media house} & \begin{tabular}[c]{@{}c@{}}\textbf{No. of}\\\textbf{ Articles}\end{tabular} & \begin{tabular}[c]{@{}c@{}}\textbf{Size on}\\\textbf {Disk}\end{tabular} \\ 
\toprule
Media House A & $40,514$                                                    & $282$M            \\ 
\midrule
Media House B & $53,024$                                                    & $364$M             \\ 
\midrule
Media House C & $31,029$                                                   & $298.6$M            \\ 
\midrule
Media House D & $27,044$                                                    & $171$M             \\ 
\bottomrule
\end{tabular}
\end{table}

\section{Person Entity Characterization with Manual Prefix Prompts}
\textit{Cloze} and \textit{Prefix} prompts are two types of prompts used as inputs for a language model to solve NLP tasks. Cloze prompts as in \cite{petroni2019language} is where the  token to be predicted is masked and the model predicts. The Prefix prompts \cite{li2021prefix, lester2021power} or the prompts used for priming when used as input to language model generates a conditional sequence text  auto-regressively.

Priming in this work can be attributed to \textit{``programming in natural language''} detailed by \citeauthor{reynolds2021prompt} \cite{reynolds2021prompt}. This work attempts to prompt language model to generate characteristics of a person entity with prompts ubiquitous in spoken and written English language. The concept is when you want to describe a person, one would express beginning with \textit{``John is described as ...''} or a semantically similar prefix, in most contexts. These prefixes and synonymous ones are very common in any corpora used to train the language models and priming with natural language phrases like \textit{``John is described as ...''} would constrain the entailment to something about \textit{John}. The intuition is prime the language model in \textit{``ubiquitous or natural language way.''} Since  these demonstrations are not very frequent in the corpus we construct a corpus of these type of sentences to fine-tune. To test this hypothesis following steps were followed with each Media House corpus and depicted in \textbf{Figure ~\ref{fig:approach}}.

\begin{enumerate}[label={\textbf{Block  \protect\threedigits{\theenumi}}:},leftmargin=*]
  \item Person Entity Mention Disambiguation in Articles
  \begin{itemize}
  \item Co-reference Replacement\footnote+{\label{neurosys}Co-reference Replacement:\\ \url{https://github.com/NeuroSYS-pl/coreference-resolution}} 
  \item Replace short names with full name
\end{itemize}
  \item First fine-tuning, GPT-2 PLM (345M) if fine-tuned  with the \textbf{Block 1} processed disambiguated articles corpus and named as \textbf{FT1 Checkpoint}  
  \item Extract clauses and their parts from sentences of person entities using \textit{spacy-clausie}\footnote+{\label{clausie}spacy-clausie: \url{https://github.com/mmxgn/spacy-clausie}} from \textbf{Block 1} disambiguated articles corpus
  \item With parts of clauses (\textbf{Block 3}) convert lemmatized verb of clauses to a gerund and construct a corpus of simple entity characterization demonstration sentences in the following pattern:\\ ``\textit{<Person\_Entity\_Name>} \textbf{`is described as'} \textit{<gerund>} \textit{<grammatically valid combination of parts of clause>}''
  From this corpus of sentences, sentences of ten entities with high frequencies in different ranges set aside as \textit{Test Corpus} and rest as \textit{Demonstrations or Training Corpus}
  \item With the \textit{Demonstrations Corpus} (\textbf{Block 4}), \textbf{FT1 Checkpoint} was fine-tuned and named as \textbf{FT2 Checkpoint}
  \item \textbf{FT2 Checkpoint} was used to generate sentences of entities in \textit{Test Corpus} with prompts defined in \textbf{Table \ref{tab:primeslist}}
  \item Sentences generated about entities in \textbf{Block 7} were tested for \textit{non-extractive characterization} against FT1 and FT2 corpus sentences using Semantic Textual Similarity \footnote+{\label{sts}STS: \url{https://www.sbert.net/docs/usage/semantic_textual_similarity.html}} and Sentiment Analysis   
\end{enumerate}
First fine-tuning to get FT1 Checkpoint is stopped \textit{at loss less than $0.6$} and Second fine-tuning to get FT2 Checkpoint is stopped \textit{at loss less than $0.1$}
\tikzstyle{blockH2} = [rectangle, draw, fill=white!20, 
text width=3cm, text centered, rounded corners, minimum height=2em]
\tikzset{box/.style={draw, minimum size=2em, text width=4.5em, text centered},
         bigbox/.style={draw, dotted, inner sep=20pt, label={north:#1},
         overlaid/.style={double copy shadow={shadow xshift=-1ex,shadow yshift=1.5ex},fill=white,draw=black,thick,minimum height = 2cm,minimum width=1cm,text width = 2cm, align=center},}
}
\begin{figure*}
\centering
\begin{tikzpicture}[
  font=\rmfamily\footnotesize,
  every matrix/.style={ampersand replacement=\&,column sep=2cm,row sep=1cm},
  source/.style={draw,thick,rounded corners,fill=yellow!20,inner sep=.3cm},
  process/.style={draw,thick,circle,fill=blue!20},
  sink/.style={source,fill=green!20},
  datastore/.style={draw,very thick,shape=datastore,inner sep=.3cm},
  dots/.style={gray,scale=2},
  to/.style={->,>=stealth',shorten >=1pt,semithick,font=\rmfamily\scriptsize},
  every node/.style={align=center}]

  \matrix{
    \node[double copy shadow={shadow xshift=-1ex,shadow yshift=1.5ex},fill=white,draw=black,thick,minimum height = 1.2cm,minimum width=1cm] (a) { Input:\\ Media House\\ Articles Corpus}; \& \node[blockH2] (e) {Block 3: Extract clauses and their parts from person entity sentences}; \& \\
    \node[blockH2] (b) {Block 1:\\ Resolve all person entity references to full names}; \& \node[blockH2] (f) {Block 4: Construct person entity characterization demonstrations}; \& \node[blockH2] (i) {Block 6:\\ Generate test entities sentences with FT2 Checkpoint };\\
    \node[double copy shadow={shadow xshift=-1ex,shadow yshift=1.5ex},fill=white,draw=black,thick,minimum height = 1.2cm,minimum width=1cm] (c) {FT1 Corpus -\\ Person Entity Mention\\  Disambiguated\\ Articles}; \& \node[double copy shadow={shadow xshift=-1ex,shadow yshift=1.5ex},fill=white,draw=black,thick,minimum height = 1.2cm,minimum width=1cm] (g) {FT2 Corpus -\\ Person Entity\\ Characterization\\ Demonstrations}; \& \node[blockH2] (j) {Block 7:\\ Test sentences against FT1 and FT2 sentences for non-extractive characterization};\\
    \node[blockH2] (d) {Block 2: Fine-tune GPT-2 PLM (345M) to create FT1 Checkpoint}; \& \node[blockH2] (h) {Block 5:\\ Fine-tune FT1 Checkpoint with demonstrations to create FT2 Checkpoint}; \& \\
  };

  \draw[to] (a) -- (b);
  \draw[to] (b) -- node[midway, left] {Output}(c);
  \draw[to] (c) -- (d);
  \draw[to] (d) -- node[midway, above] {FT1 Checkpoint} (h);
  \draw[to] (c.east) to[bend left=14] (e.west);
  \draw[to] (e) -- (f);
  \draw[to] (f) -- node[midway, left] {Output} (g);
  \draw[to] (g) -- (h);
  \draw[to] (h.east) to[bend left=16] node[midway, below]  {FT2 Checkpoint}(i.west);
  \draw[to] (i) -- node[midway, above]{Generated\hspace{0.1cm}Sentences} (j);
  \node[bigbox=First Fine-tuning, fit=(a) (b) (c) (d),] (FT1) {};
  \node[bigbox=Second Fine-tuning,fit=(e) (f) (g) (h),] (FT2) {};
  \node[bigbox=Validation,fit=(i) (j)] (IJ) {};
\end{tikzpicture}
\caption{Pipeline of processing a Media House Corpus, generating sentences about entities and validating for characterizations. \textit{Block 1} uses NeuroSYS\tablefootnote{\ref{neurosys}}. 
 \textit{Block 3} uses Claucy\tablefootnote{\ref{clausie}}. Details of FT2 or Demonstrations Corpus is shown in Table ~\ref{tab:ftdataset}. In 
 \textit{Block 6} sentences about test entities are generated with prompts listed in Table ~\ref{tab:primeslist}. \textit{Block 7} uses Semantic Text Similarity (STS)\tablefootnote{\ref{sts}} to compare generated sentence with corpus sentences. Examples of generated and semantically similar corpus sentences are shown in Table ~\ref{tab:tp_examples}}
\label{fig:approach}
\end{figure*}
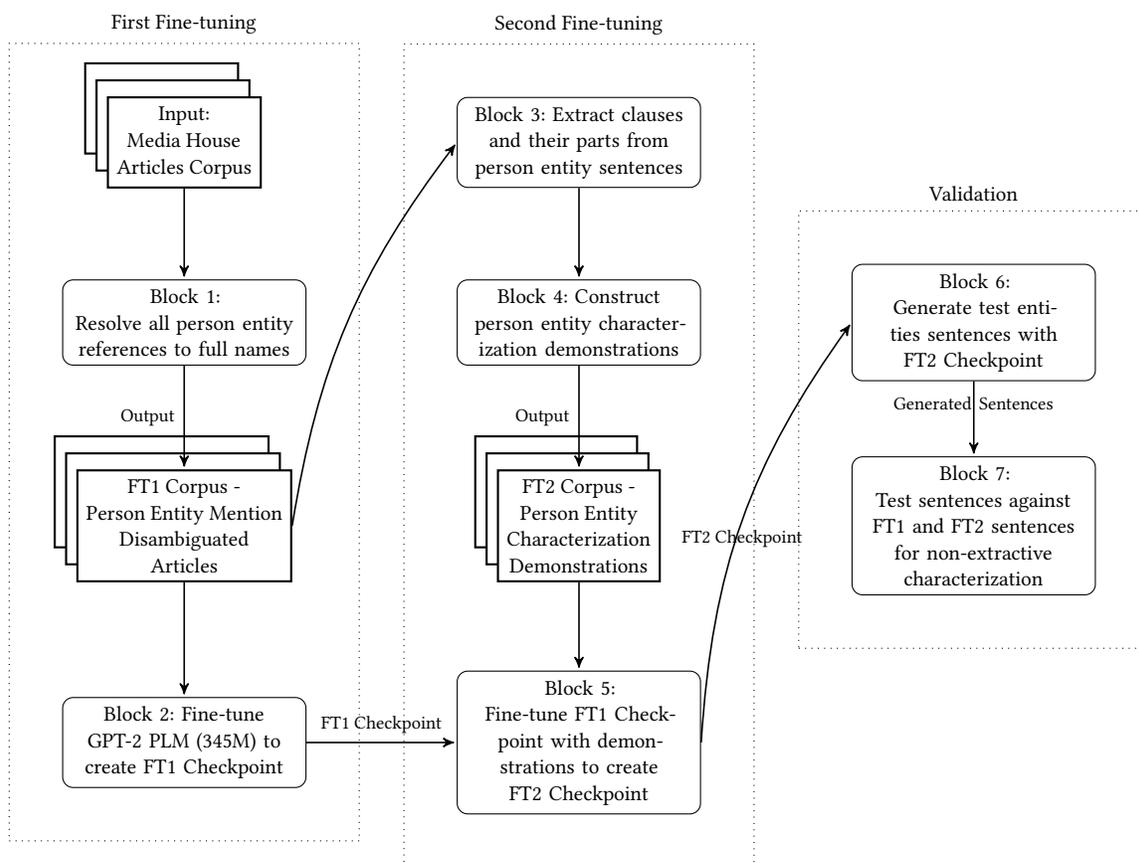

Following subsections detail each of the above steps.
\begin{table}[!htbp]
\centering
\caption{Four types of \textit{prefix prompts} used to \\ generate sentences about entities}
\label{tab:primeslist}
\begin{tabular}{c} 
\toprule
``\textit{<Person\_Entity\_Name>} is described as \\being''   \\
\midrule
``\textit{<Person\_Entity\_Name>} is described as \\having characteristics''    \\
\midrule
``\textit{<Person\_Entity\_Name>} is described as \\performing''    \\
\midrule
``\textit{<Person\_Entity\_Name>} is described as \\stating''      \\ 
\bottomrule
\end{tabular}
\end{table}

\subsection{Person Entity Mention Disambiguation}
Co-reference resolution improves the accuracy of NLP tasks like machine translation, sentiment analysis, paraphrase detection and summarization (\citeauthor{sukthanker2020anaphora}) \cite{sukthanker2020anaphora}. We have disambiguated person entity mentions in the articles to ensure that every person entity sentence has full name of the entity.

The first pre-processing was replacing entity co-references with the actual entity name and on this output replace partial name references with full name to finally get a processed document with full name of the entity in maximum number of sentences in each news article.

NeuroSYS coreference-resolution\footnote{\ref{neurosys}} proposes three intersection strategies or ensemble methods of AllenNLP and Huggingface coreference models outputs. The methods are \textit{strict} where clusters identical in both the models are considered, \textit{partial} where spans identical in both model outputs are considered and \textit{fuzzy} where spans and overlapping spans are considered from both the models. In this work we leveraged the \textit{fuzzy} ensemble and processed the raw articles.

The objective of this work was to generate single concise sentences of person entity characterizations. To align with this objective the sentences in each media house articles were processed to contain unambiguous entity mentions. To address this, the co-references replaced texts were processed to replace partial name references with full name of the entity so that every sentence has full qualified mention of the entity and information about the entity. To achieve this, we followed a logic of processing one article at a time, mapping partial names, either first name or last name, with the full name by comparing tokens. The intuition is that entity is referred with full name in the initial parts of the article and in later sentences of the article either first name or last name is used to refer to the entity. The partial name should be either first name or last name of the entity in the previously used longer name. 

This final corpus of articles with full entity disambiguation was used for first fine tuning or FT1. For all the media houses the loss plateaued around 0.6 and hence a checkpoint around this loss was considered for next fine tuning FT2. 

\subsection{Characterization Sentences Corpus of Person Entities}
To generate simple concise demonstration sentences of entity characterizations, the FT1 Checkpoint was fine-tuned with manual prompt prefixed to clauses of entities. Clauses contain the main information of entities. Corpus of simple sentences of anything said, done or events related to the entities was constructed using clauses and their parts extracted from each article using ClauCy\footnote{\ref{clausie}} ~\cite{del2013clausie}. Clauses and their parts were extracted from each sentence in articles. The parts of the clauses are: \textit{Type, Subject, Verb, Indirect\_Object, Direct\_Object, Complement and Adverbials}. There are ten Clause Types with combination of parts: \textit{SVC, SVOO, SVOC, SVO, SVOA, SVO, SVO, SV, SVA and SVO}. Every clause has a subject and verb, other parts vary depending on the input sentence. Entities and the sentences they appear were mapped and maps with more than $500$ sentences were considered for FT2 corpus. Table \ref{tab:ftdataset} shows the details of FT2 sentences corpus for each media house.
\begin{table}[]
\centering
\scriptsize
\begin{tabular}{@{}lcccc@{}}
\toprule
Clause Type &
  \begin{tabular}[c]{@{}c@{}}Media \\ House 1\end{tabular} &
  \begin{tabular}[c]{@{}c@{}}Media\\ House 2\end{tabular} &
  \begin{tabular}[c]{@{}c@{}}Media\\ House 3\end{tabular} &
  \begin{tabular}[c]{@{}c@{}}Media\\ House 4\end{tabular} \\ \midrule
SV                                                                    & 11,349 & 3,244  & 17,863   & 7,768  \\ \midrule
SVA                                                                   & 2,732  & 695  & 3,829    & 1,698  \\ \midrule
SVC                                                                   & 26,042 & 10,403  & 40,899   & 16,617 \\ \midrule
SVO                                                                   & 23,522 & 8,750  & 34,164   & 15,795 \\ \midrule
SVOA                                                                  & 1,223  & 488  & 1,915    & 937    \\ \midrule
SVOC                                                                  & 2,832  & 1,249  & 4,619    & 1,857  \\ \midrule
SVOO                                                                  & 597    & 246  & 738      & 370    \\ \toprule \\ \toprule
\begin{tabular}[c]{@{}l@{}}FT2 Dataset\\ Sentences Count\end{tabular} & 68,297 & 25,075  & 1,04,027 & 45,042 \\ \midrule
\begin{tabular}[c]{@{}l@{}}Unique Person\\Entities Count\end{tabular}      & 117    & 69 & 140      & 83     \\ \bottomrule
\end{tabular}\caption{Each Media House FT2 Sentences Corpus details. Count of each extracted clause type, total number of sentences and unique person entities in each corpus}
\label{tab:ftdataset}
\end{table}

FT2 sentences were constructed by suffixing Subject with ``is described as'', converting Verb in to Gerund form and grammatically joining other parts of the clause to form a complete readable sentence. Gerund or present participle is the adjective form the verb (like showing, saying, claiming, winning, etc.) and functions as attributing the other parts of the clause (Object, Complements and Adverbials) to the Subject. Ten subjects or person entities with highest count in different ranges were separated as test corpus and rest of entity sentences for second fine tuning. This was done to ensure testing with entity count in broad ranges. The checkpoint from FT1 was further fine tuned with FT2 corpus. For all the media houses, the second fine tuning plateaued around loss of 0.1 and hence fine tuning was stopped when loss reached below 0.1.

\subsection{Generative Entity Characterization}
Widely prevalent manual prompts in the spoken and written language used to talk about a person were chosen to prime the language model. Sentences were generated with the FT2 Checkpoint. The second fine-tuning, FT2, was with a corpus of sentences with ``is described as'' prompt. The results of the generated sentences with this prompt were not convincing, so we experimented with semantic alternative prompts shown in Table \ref{tab:primeslist}. With these prompts, we observed entity characterizing generated sentences. Ideally, all the test sentences should be generated; hence, sentences were generated to each entity's count in the test corpus. Novel combinations of information in the corpus or summarized opinions of test entities were expected in the generated texts. The generated texts were compared for Semantic Textual Similarity (STS) with FT1 and FT2 corpus sentences using Sentence Transformers\footnote{\ref{sts}}. Since language models are probabilistic and generate novel sentences, we chose cosine similarity of greater than or equal to 0.6 as a positive result.

To the best of our knowledge, there is no start-of-the-art corpus for Entity Characterization demonstrations and evaluation criteria. For this purpose, we have compiled FT2 dataset and defined evaluation criteria with Confusion Matrix as shown in Table ~\ref{tab:cm}. 

\begin{table*}[!htbp]
\centering
\scriptsize
\begin{tabular}{@{}l|l@{}}
\toprule
\multicolumn{1}{c}{\textbf{True Positive (TP)}}                  & \multicolumn{1}{c}{\textbf{Type 2 Error or False Positive (FP)}} \\ \midrule
\begin{tabular}[c]{@{}l@{}}\textbf{Definition:} \textbf{\textit{Novel and Meaningful or Non-extractive Characterization.}} \\ The generated sentence has a high semantically matching sentence \\ in FT1 or FT2 datasets, and the person entity in both sentence\\ contexts are the same.\\ \textbf{Condition:}\\   Prompt Entity == Ground Truth Entity\\    And\\    The cosine score between generated sentence and FT1 or FT2\\     dataset sentence   is \textgreater{}= 0.6\end{tabular} &
  \begin{tabular}[c]{@{}l@{}}\textbf{Definition:} The generated sentence has a high semantically\\ matching sentence in FT1 or FT2   datasets and the person entity in\\ both sentence contexts are different.\\ \textbf{Condition:}\\   Prompt Entity != Ground Truth Entity\\    And\\    The cosine score between generated sentence and FT1 or FT2\\    dataset sentence   is \textgreater{}= 0.6\end{tabular} \\ \midrule
\multicolumn{1}{c}{\textbf{Type 1 Error or False Negative (FN)}} & \multicolumn{1}{c}{\textbf{True   Negative (TN)}}                \\ \midrule
\begin{tabular}[c]{@{}l@{}}\textbf{Definition:} The generated sentence has a low semantically\\ matching sentence in FT1 or FT2 datasets, and the person entity in \\ both sentence contexts are the same.\\ \textbf{Condition:}\\   Prompt Entity == Ground Truth Entity\\    And\\    The cosine score between generated sentence and FT1 or FT2\\    dataset sentence   is \textless 0.6\end{tabular} &
  \begin{tabular}[c]{@{}l@{}}\textbf{Definition:}   The generated sentence has a low semantically\\ matching sentence in FT1 or FT2   datasets and the person entity in\\ both sentence contexts are different.\\ \textbf{Condition:}\\   Prompt Entity != Ground Truth Entity\\   And\\   The cosine score between generated sentence and FT1 or FT2\\   dataset sentence   is \textless 0.6\end{tabular} \\ \bottomrule
\end{tabular}
\caption{Person Entity Characterization Evaluation Criteria}
\label{tab:cm}
\end{table*}

\begin{table*}[!htbp]
\scriptsize
\centering
\begin{tabular}{@{}cccccccccccc@{}}
\toprule
\multirow{2}{*}{Manual Prompts}          & \multicolumn{1}{c}{\multirow{2}{*}{\begin{tabular}[c]{@{}c@{}}Distinct\\ Generated\\ Sentences\\ Count\end{tabular}}} & \multicolumn{2}{c}{\begin{tabular}[c]{@{}c@{}}\%\\ of\\ Distinct\\ Semantic\\ Matches\end{tabular}} & \multicolumn{2}{c}{\begin{tabular}[c]{@{}c@{}}Average \\ Sentiment Scores \\ Differences of \\ True Positives (TP)\end{tabular}} & \multicolumn{2}{c}{\begin{tabular}[c]{@{}c@{}}F1\\ Score\end{tabular}} & \multicolumn{2}{c}{Precision}                          & \multicolumn{2}{c}{Recall}        \\ \cmidrule(l){3-12} 
                                         & \multicolumn{1}{c}{}                                                                                                  & \multicolumn{1}{c}{FT1}                          & \multicolumn{1}{c}{FT2}                         & \multicolumn{1}{c}{FT1}                                          & \multicolumn{1}{c}{FT2}                                         & \multicolumn{1}{c}{FT1}           & \multicolumn{1}{c}{FT2}           & \multicolumn{1}{c}{FT1}   & \multicolumn{1}{c}{FT2}   & \multicolumn{1}{c}{FT1}   & FT2   \\ \midrule
\multicolumn{1}{l}{}                   & \multicolumn{11}{c}{\textbf{Media   House 1}}                                                                                                                                                                                                                                                                                                                                                                                                                                                                                                         \\ \midrule
\textbf{\textit{is described   as having characteristics}} & \multicolumn{1}{c}{3010}                                                                                              & \multicolumn{1}{c}{41\%}                         & \multicolumn{1}{c}{28\%}                        & \multicolumn{1}{c}{0.157}                                        & \multicolumn{1}{c}{0.108}                                       & \multicolumn{1}{c}{\textbf{\underline{0.864}}}         & \multicolumn{1}{c}{0.54}          & \multicolumn{1}{c}{\textbf{\underline{0.888}}} & \multicolumn{1}{c}{0.504} & \multicolumn{1}{c}{\textbf{\underline{0.842}}} & 0.981 \\ \midrule
\textbf{\textit{is described   as being}}                  & \multicolumn{1}{c}{7347}                                                                                              & \multicolumn{1}{c}{50\%}                         & \multicolumn{1}{c}{34\%}                        & \multicolumn{1}{c}{0.154}                                        & \multicolumn{1}{c}{0.136}                                       & \multicolumn{1}{c}{\textbf{\underline{0.898}}}         & \multicolumn{1}{c}{0.497}         & \multicolumn{1}{c}{\textbf{\underline{0.852}}} & \multicolumn{1}{c}{0.414} & \multicolumn{1}{c}{\textbf{\underline{0.948}}} & 0.973 \\ \midrule
is described   as performing             & \multicolumn{1}{c}{4243}                                                                                              & \multicolumn{1}{c}{28\%}                         & \multicolumn{1}{c}{20\%}                        & \multicolumn{1}{c}{0.034}                                        & \multicolumn{1}{c}{0.027}                                       & \multicolumn{1}{c}{0.725}         & \multicolumn{1}{c}{0.317}         & \multicolumn{1}{c}{0.607} & \multicolumn{1}{c}{0.294} & \multicolumn{1}{c}{0.899} & 0.968 \\ \midrule
is described   as stating                & \multicolumn{1}{c}{8901}                                                                                              & \multicolumn{1}{c}{65\%}                         & \multicolumn{1}{c}{44\%}                        & \multicolumn{1}{c}{0.138}                                        & \multicolumn{1}{c}{0.097}                                       & \multicolumn{1}{c}{0.807}         & \multicolumn{1}{c}{0.406}         & \multicolumn{1}{c}{0.726} & \multicolumn{1}{c}{0.363} & \multicolumn{1}{c}{0.907} & 0.935 \\ \midrule
                                         & \multicolumn{11}{c}{\textbf{Media   House 2}}                                                                                                                                                                                                                                                                                                                                                                                                                                                                                                         \\ \midrule
\textbf{\textit{is described   as having characteristics}} & \multicolumn{1}{c}{4407}                                                                                              & \multicolumn{1}{c}{27\%}                         & \multicolumn{1}{c}{17\%}                        & \multicolumn{1}{c}{0.062}                                        & \multicolumn{1}{c}{0.054}                                       & \multicolumn{1}{c}{\textbf{\underline{0.910}}}          & \multicolumn{1}{c}{0.55}          & \multicolumn{1}{c}{\textbf{\underline{0.892}}} & \multicolumn{1}{c}{0.622} & \multicolumn{1}{c}{\textbf{\underline{0.929}}} & 0.492 \\ \midrule
\textbf{\textit{is described   as being}}                  & \multicolumn{1}{c}{4985}                                                                                              & \multicolumn{1}{c}{51\%}                         & \multicolumn{1}{c}{32\%}                        & \multicolumn{1}{c}{0.155}                                        & \multicolumn{1}{c}{0.139}                                       & \multicolumn{1}{c}{\textbf{\underline{0.894}}}         & \multicolumn{1}{c}{0.54}          & \multicolumn{1}{c}{\textbf{\underline{0.837}}} & \multicolumn{1}{c}{0.519} & \multicolumn{1}{c}{\textbf{\underline{0.960}}} & 0.563 \\ \midrule
is described   as stating                & \multicolumn{1}{c}{5794}                                                                                              & \multicolumn{1}{c}{67\%}                         & \multicolumn{1}{c}{41\%}                        & \multicolumn{1}{c}{0.126}                                        & \multicolumn{1}{c}{0.099}                                       & \multicolumn{1}{c}{0.825}         & \multicolumn{1}{c}{0.469}         & \multicolumn{1}{c}{0.734} & \multicolumn{1}{c}{0.476} & \multicolumn{1}{c}{0.942} & 0.461 \\ \midrule
is described   as performing             & \multicolumn{1}{c}{2506}                                                                                              & \multicolumn{1}{c}{30\%}                         & \multicolumn{1}{c}{20\%}                        & \multicolumn{1}{c}{0.023}                                        & \multicolumn{1}{c}{0.016}                                       & \multicolumn{1}{c}{0.557}         & \multicolumn{1}{c}{0.263}         & \multicolumn{1}{c}{0.404} & \multicolumn{1}{c}{0.184} & \multicolumn{1}{c}{0.898} & 0.467 \\ \midrule
                                         & \multicolumn{11}{c}{\textbf{Media   House 3}}                                                                                                                                                                                                                                                                                                                                                                                                                                                                                                         \\ \midrule
\textbf{\textit{is described   as having characteristics}} & \multicolumn{1}{c}{5418}                                                                                              & \multicolumn{1}{c}{22\%}                         & \multicolumn{1}{c}{20\%}                        & \multicolumn{1}{c}{0.102}                                        & \multicolumn{1}{c}{0.079}                                       & \multicolumn{1}{c}{\textbf{\underline{0.953}}}         & \multicolumn{1}{c}{\textbf{\underline{0.743}}}         & \multicolumn{1}{c}{0.945} & \multicolumn{1}{c}{\textbf{\underline{0.682}}} & \multicolumn{1}{c}{\textbf{\underline{0.960}}} & \textbf{\underline{0.816}} \\ \midrule
\textbf{\textit{is described   as being}}                  & \multicolumn{1}{c}{9591}                                                                                              & \multicolumn{1}{c}{47\%}                         & \multicolumn{1}{c}{59\%}                        & \multicolumn{1}{c}{0.177}                                        & \multicolumn{1}{c}{0.142}                                       & \multicolumn{1}{c}{\textbf{\underline{0.921}}}         & \multicolumn{1}{c}{0.597}         & \multicolumn{1}{c}{\textbf{\underline{0.889}}} & \multicolumn{1}{c}{0.525} & \multicolumn{1}{c}{\textbf{\underline{0.954}}} & 0.692 \\ \midrule
is described   as performing             & \multicolumn{1}{c}{6430}                                                                                              & \multicolumn{1}{c}{35\%}                         & \multicolumn{1}{c}{39\%}                        & \multicolumn{1}{c}{0.064}                                        & \multicolumn{1}{c}{0.039}                                       & \multicolumn{1}{c}{0.869}         & \multicolumn{1}{c}{0.576}         & \multicolumn{1}{c}{0.828} & \multicolumn{1}{c}{0.517} & \multicolumn{1}{c}{0.915} & 0.650 \\ \midrule
is described   as stating                & \multicolumn{1}{c}{11222}                                                                                             & \multicolumn{1}{c}{59\%}                         & \multicolumn{1}{c}{30\%}                        & \multicolumn{1}{c}{0.150}                                        & \multicolumn{1}{c}{0.117}                                       & \multicolumn{1}{c}{0.844}         & \multicolumn{1}{c}{0.515}         & \multicolumn{1}{c}{0.767} & \multicolumn{1}{c}{0.465} & \multicolumn{1}{c}{0.940} & 0.576 \\ \midrule
                                         & \multicolumn{11}{c}{\textbf{Media   House 4}}                                                                                                                                                                                                                                                                                                                                                                                                                                                                                                         \\ \midrule
\textbf{\textit{is described   as having characteristics}} & \multicolumn{1}{c}{177}                                                                                               & \multicolumn{1}{c}{42\%}                         & \multicolumn{1}{c}{23\%}                        & \multicolumn{1}{c}{0.024}                                        & \multicolumn{1}{c}{0.038}                                       & \multicolumn{1}{c}{0.789}         & \multicolumn{1}{c}{\textbf{\underline{0.824}}}         & \multicolumn{1}{c}{0.679} & \multicolumn{1}{c}{\textbf{\underline{0.860}}} & \multicolumn{1}{c}{0.942} & \textbf{\underline{0.791}} \\ \midrule
is described   as performing             & \multicolumn{1}{c}{4478}                                                                                              & \multicolumn{1}{c}{29\%}                         & \multicolumn{1}{c}{20\%}                        & \multicolumn{1}{c}{0.025}                                        & \multicolumn{1}{c}{0.011}                                       & \multicolumn{1}{c}{0.754}         & \multicolumn{1}{c}{0.622}         & \multicolumn{1}{c}{0.660} & \multicolumn{1}{c}{0.638} & \multicolumn{1}{c}{0.879} & 0.607 \\ \midrule
\textbf{\textit{is described   as being}}                  & \multicolumn{1}{c}{5375}                                                                                              & \multicolumn{1}{c}{48\%}                         & \multicolumn{1}{c}{32\%}                        & \multicolumn{1}{c}{0.156}                                        & \multicolumn{1}{c}{0.110}                                       & \multicolumn{1}{c}{\textbf{\underline{0.903}}}         & \multicolumn{1}{c}{0.574}         & \multicolumn{1}{c}{\textbf{\underline{0.874}}} & \multicolumn{1}{c}{0.548} & \multicolumn{1}{c}{\textbf{\underline{0.934}}} & 0.601 \\ \midrule
is described   as stating                & \multicolumn{1}{c}{6420}                                                                                              & \multicolumn{1}{c}{60\%}                         & \multicolumn{1}{c}{39\%}                        & \multicolumn{1}{c}{0.139}                                        & \multicolumn{1}{c}{0.090}                                       & \multicolumn{1}{c}{0.837}         & \multicolumn{1}{c}{0.464}         & \multicolumn{1}{c}{0.789} & \multicolumn{1}{c}{0.476} & \multicolumn{1}{c}{0.892} & 0.452 \\ \bottomrule
\end{tabular}
\caption{Metrics based on evaluation criteria in \textbf{Table ~\ref{tab:cm}} of FT2 checkpoint \\ generated sentences with FT1 and FT2 corpus sentences}
\label{tab:resultsall}
\end{table*}
\begin{table*}[!htbp]
\centering
\tiny
\begin{tabular}{@{}ll@{}}
\toprule
\multicolumn{2}{c}{\textbf{Examples of Novel and Meaningful or Non-extractive Person Entity Characterizations (True Positives)}} \\ \midrule
\multicolumn{1}{c}{\textbf{Generated Text}} &
  \multicolumn{1}{c}{\textbf{Corpus   Text}} \\ \midrule
\multicolumn{2}{c}{\textbf{Media House 1 - FT1}} \\ \midrule
\multicolumn{1}{l}{Entity A is described as having characteristics that can end up   forming the government in State.} &
  As   per sources, Entity A is tipped to be the next Chief of Elected Memebers of   State. \\ \midrule
\multicolumn{1}{l}{Entity B is described as having characteristics like   threatening, stoking violence, etc.} &
  Entity   B comments come after he was likened to a terrorist by a prominent leader. \\ \midrule
\multicolumn{1}{l}{Entity C is described as having characteristics of a caring   truly, a loving truly, and a pious truly} &
  Entity   C was a great leader with a great sense of compassion and humour. \\ \midrule
\multicolumn{1}{l}{Entity D is described as having the characteristics of an Angel.} &
  \begin{tabular}[c]{@{}l@{}}A   prominent chronicles of a Powerful person, a character that bears an uncanny\\   resemblance to Entity D.\end{tabular} \\ \midrule
\multicolumn{1}{l}{Entity E is described as being critical of the Prominent Party   government in state.} &
  Leader   Entity E had remained highly critical of the Prominent Party government in   the past. \\ \midrule
\multicolumn{1}{l}{Entity D is described as being a strong advocate for the   interests of the people.} &
  \begin{tabular}[c]{@{}l@{}}Listing   out the various pro-people initiatives launched by Entity D, a Rebel leader\\  said Entity D was a popular leader who always remained in the hearts of the   people.\end{tabular} \\ \midrule
\multicolumn{1}{l}{Entity F is described as being an extremely beautiful face} &
  The   smoky eyes and nude lips further complimented Entity F's look. \\ \midrule
\multicolumn{1}{l}{Entity G is described as being very quick in taking the   decision, in such a situation.} &
  Entity   G, however, is the first politician from the ruler's family to have reacted   to the step. \\ \midrule
\multicolumn{2}{c}{\textbf{Media House 2 - FT1}} \\ \midrule
\multicolumn{1}{l}{Entity H is described as being under house arrest,  at his residence.} &
  Does   this mean party head and elected member Entity H is under house arrest? \\ \midrule
\multicolumn{1}{l}{Entity I is described as being the new Chief of State.} &
  "People   of State want Entity I to become the Chief of State," he added. \\ \midrule
\multicolumn{1}{l}{Entity H is described as being unwell.} &
  "We   have heard that Entity H is unwell, which is understandable" \\ \midrule
\multicolumn{1}{l}{Entity J is described as being mature.} &
  \begin{tabular}[c]{@{}l@{}}Another   significant development is that Entity J has emerged as a matured leader\\ during the General elections.\end{tabular} \\ \midrule
\multicolumn{1}{l}{\begin{tabular}[c]{@{}l@{}}Entity J is described as having the characteristics like a true leader and\\ a person who has a capacity to thought.\end{tabular}} &
  \multicolumn{1}{l}{
  He   said that Entity J has a good vision and thoughts.} \\ \midrule
\multicolumn{1}{l}{\begin{tabular}[c]{@{}l@{}}Entity K is described as having characteristics such as being able\\ to bring people on the path of development.\end{tabular}} &
  \begin{tabular}[c]{@{}l@{}}"Through   better economic management, we could take the common man ahead on the\\ path of progress," Entity K claimed.\end{tabular} \\ \midrule
\multicolumn{1}{l}{\begin{tabular}[c]{@{}l@{}}Entity L is described as having characteristics like a seasoned   politician and\\ leader and an ideal organisational person.\end{tabular}} &
  \begin{tabular}[c]{@{}l@{}}Entity   L is an extremely qualified \& respected leader, Entity L has served this   nation with \\ dedication \& humility.\end{tabular} \\ \midrule
\multicolumn{1}{l}{\begin{tabular}[c]{@{}l@{}}Entity M is described as having characteristics such as   reconciling to the family,\\ developing friendships that helped him during the   difficult times, honesty and integrity\\ in discharge of his duties as an   actor.\end{tabular}} &
  \begin{tabular}[c]{@{}l@{}}She   said, "actor Entity M has really had my back, and has been there for me   as a friend and\\ support over the years, unfailingly and intuitively."\end{tabular} \\ \midrule
\multicolumn{1}{l}{Entity N is described as being no entry, in the roadshow.} & \begin{tabular}[c]{@{}l@{}}
  Entity   N said that he was restricted only to his region as he does not hold any   official post\\ in city unit.\end{tabular} \\ \midrule
\multicolumn{1}{l}{Entity M is described as being an awareness campaign to urge   people to follow.} & \begin{tabular}[c]{@{}l@{}}
  During   this time, Entity M has appeared in several public safety videos, urging his   fans to\\ obey laws.\end{tabular} \\ \midrule
\multicolumn{2}{c}{\textbf{Media House 3 - FT1}} \\ \midrule
\multicolumn{1}{l}{Entity D is described as having characteristics of a strong   personality.} &
  On   one side, you see in Entity D a woman who was the personification of   authoritarianism. \\ \midrule
\multicolumn{1}{l}{Entity O is described as having characteristics of a classic   leader born to influential parents.} &
  \begin{tabular}[c]{@{}l@{}}With   a massive campaign focused on Entity O's personality, he has towered over   other\\ stalwarts in State politics, including a Top Leader and his father's   father.\end{tabular} \\ \midrule
\multicolumn{1}{l}{Entity J is described as being to become the President.} &
  Entity   J finally looks all set to become President. \\ \midrule
\multicolumn{1}{l}{Entity P is described as being the primary link between the   party and the people} & \begin{tabular}[c]{@{}l@{}}
  "Entity   P is the unifying factor for party," the party affairs representative   told\\ in an interview.\end{tabular} \\ \midrule
\multicolumn{2}{c}{\textbf{Media House 3 - FT2}} \\ \midrule
\multicolumn{1}{l}{\begin{tabular}[c]{@{}l@{}}Entity P is described as having characteristics of a leader who   has a habit of wearing\\ her aspirational state's uniforms.\end{tabular}} &
  Entity   P is described as coming in her uniform. \\ \midrule
\multicolumn{1}{l}{Entity E is described as having characteristics of a leader who   may be able to win City elections.} &
  \begin{tabular}[c]{@{}l@{}}Entity   E is described as claiming he built his from the ground up by addressing   dozens of\\ rallies in State's villages and towns, before converging in City.\end{tabular} \\ \midrule
\multicolumn{1}{l}{Entity Q is described as having characteristics of a successful   orator.} &
  Entity   Q is described as making that comment , in his personal capacity. \\ \midrule
\multicolumn{1}{l}{\begin{tabular}[c]{@{}l@{}}Entity R is described as having characteristics of a leader who   may need to rein\\ in elements on the ground.\end{tabular}} &
  \begin{tabular}[c]{@{}l@{}}Entity   R is described as saying that he will take all efforts to help authorities \\ contain the spread of the disease.\end{tabular} \\ \midrule
\multicolumn{2}{c}{\textbf{Media House 4 - FT1}} \\ \midrule
\multicolumn{1}{l}{Entity B is described as being in State,  for a two-day visit to State.} &
  Entity   B is on a two-day visit to State. \\ \midrule
\multicolumn{1}{l}{Entity S is described as being active, on social media.} &
  Entity   S is an avid social media player and also a writes a blog regularly. \\ \midrule
\multicolumn{1}{l}{Entity T is described as being the new go-to girl.} &
  New   'Country Girl' Entity T is making a lot of headlines these days. \\ \midrule
\multicolumn{1}{l}{Entity U is described as being in no mood to waste time.} &
  "I   do not waste my time on what he says," said the leader Entity U. \\ \midrule
\multicolumn{2}{c}{\textbf{Media House 4 - FT2}} \\ \midrule
\multicolumn{1}{l}{Entity J is described as having characteristics of a   revolutionary.} &
  Entity   J is described as showing hiss mettle. \\ \midrule
\multicolumn{1}{l}{Entity V is described as having characteristics of an artiste.} &
  \begin{tabular}[c]{@{}l@{}}Entity   V is described as winning several accolades for his work, including the\\ Country Award for his debut role as a child artist.\end{tabular} \\ \midrule
\multicolumn{1}{l}{Entity W is described as having characteristics of a leader.} &
  Entity   W is described as charting his future course of action. \\ \bottomrule
\end{tabular}
\caption{True Positive examples of top metrics in \textbf{Table ~\ref{tab:resultsall}}}
\label{tab:tp_examples}
\end{table*}

\begin{table*}[!htbp]
\centering
\scriptsize
\begin{tabular}{@{}cl@{}}
\toprule
\textbf{Test   Entity} &
  \textbf{Examples of   Generated Characterizations Across Media Houses} \\ \midrule
\multirow{8}{*}{\textbf{Entity   1}} &
  \multirow{2}{*}{\begin{tabular}[c]{@{}l@{}}MH1: \textit{is described as having characteristics} of an immature,  perhaps naive,  leader\\ MH1: \textit{is described as having characteristics} of an immature,  perhaps anti-national,  protestor\end{tabular}} \\
 &
   \\ \cmidrule(l){2-2} 
 &
  \multirow{2}{*}{\begin{tabular}[c]{@{}l@{}}MH2: \textit{is described as having characteristics} like a true   patriot\\ MH2: \textit{is described as having characteristics} like a true leader and a man to trust\end{tabular}} \\
 &
   \\ \cmidrule(l){2-2} 
 &
  \multirow{2}{*}{\begin{tabular}[c]{@{}l@{}}MH3: \textit{is described as having characteristics} of a classic Party loyalist\\ MH3: \textit{is described as having characteristics} of a leader who is adept at top command\end{tabular}} \\
 &
   \\ \cmidrule(l){2-2} 
 &
  \multirow{2}{*}{\begin{tabular}[c]{@{}l@{}}MH4: \textit{is described as being} at loggerheads with the Party leadership\\ MH4: \textit{is described as being} fit, also,  to be a prime minister\end{tabular}} \\
 &
   \\ \midrule
\multirow{6}{*}{\textbf{Entity   2}} &
  \multirow{2}{*}{\begin{tabular}[c]{@{}l@{}}MH1: \textit{is described as having characteristics} of a strong woman\\ MH1: \textit{is described as having characteristics} of a strong political personality\end{tabular}} \\
 &
   \\ \cmidrule(l){2-2} 
 &
  \multirow{2}{*}{\begin{tabular}[c]{@{}l@{}}MH2: \textit{is described as having characteristics} such as long history with the State and its unique culture and languages\\ MH2: \textit{is described as having characteristics} like a person, strong willpower, and political instincts\end{tabular}} \\
 &
   \\ \cmidrule(l){2-2} 
 &
  \multirow{2}{*}{\begin{tabular}[c]{@{}l@{}}MH3: \textit{is described as having characteristics} of a classic leader\\ MH3: \textit{is described as having characteristics} of a strong regional leader\end{tabular}} \\
 &
   \\ \midrule
\multirow{4}{*}{\textbf{Entity   3}} &
  \multirow{2}{*}{\begin{tabular}[c]{@{}l@{}}MH3: \textit{is described as having characteristics} of a leader who is adept at stoking   passions through the Party's various programs\\ MH3: \textit{is described as having characteristics} like a leader with firm control over the party, a decisive figure, and an ability to move the front\end{tabular}} \\
 &
   \\ \cmidrule(l){2-2} 
 &
  \multirow{2}{*}{\begin{tabular}[c]{@{}l@{}}MH4: \textit{is described as being} successful in expanding the Party\\ MH4: \textit{is described as being} a "prominent face" of the Party\end{tabular}} \\
 &
   \\ \bottomrule
\end{tabular}
\caption{Examples of Entity Characterizations across media houses}
\label{tab:entitybymh}
\end{table*}
The following section details the results of entity characterizations generated with prefix prompts in Table \ref{tab:primeslist} 

\section{Results}
With the FT2 checkpoint of each media house, sentences were generated for ten test entities, with four prompts shown in Table \ref{tab:primeslist}, to test the hypothesis. The count of generated sentences was up to the entity sentences count in the FT2 sentences corpus. The length of the generated text was limited to 30, and the first sentence in the generated text was considered for evaluation.
The first evaluation was with the FT2 sentences corpus. Entity names in the FT2 sentences corpus was masked and embeddings were constructed. Then, each generated sentence matched with all sentence embeddings. Masking entity names in FT2 corpus resulted in better relevant matches. The match with the highest cosine score was considered the best semantic match. Next, a similar evaluation was done with the FT1 articles corpus. Every sentence was extracted from each article of the FT1 corpus, and sentences with person entities and lengths greater than ten were considered to compare with the generated text to consider sentences with reasonable information content and to exclude insignificant sentences. In this evaluation entity names were not masked in the FT1 corpus sentences.

We define the evaluation criteria as detailed in Table \ref{tab:cm}. The evaluation approach is that if a generated sentence is semantically similar to an FT1 or FT2 sentence, the entity referred to is the same. Then the generated sentence should be about the entity. In FT2, we have processed sentences where something said, done, and about an event related to the entity is suffixed with entity name and "is described as" and we refer to these sentences as the entity characterizing sentences. The FT2 generated sentences were the same kind as in the FT2 corpus. Examples in Table ~\ref{tab:tp_examples}.
Hence we define the FT2 generated sentences as characterizations and validate the characterizations with the Confusion Matrix definitions in Table \ref{tab:cm}. Also, good metrics on either FT1 or FT2 dataset is good enough to conclude soundness of the approach.

Table ~\ref{tab:resultsall} shows the metrics derived from the evaluation criteria. F1, Precision and Recall are  computed based on the \textit{Distinct Generated Sentences Count}. was shown, the \textit{``is described as having characteristics''} and \textit{``is described as being''} prompts resulted in good F1, Precision, and Recall (or True Positive Rate) scores across media houses, which is Confirming that FT2 would lead to generating the most relevant sentences to the entity. More than one generated sentence is semantically similar to a corpus sentence. For True Positives average of the difference in sentiment scores of generated and semantically matching sentence is marginal. Therefore, it is encouraging to conclude that FT2 generated sentences are about the prompted entities and characterizing the entities with sentiment in the corpus. An exhaustive examples of generated True Positive and corresponding semantically matching sentences of top metrics in Table ~\ref{tab:resultsall} is shown in Table ~\ref{tab:tp_examples}.

With the approach, evaluation criteria, and test prompts detailed in this work, the \textit{``is described as having characteristics''} and \textit{``is described as being''} manual prompts function reasonably well as prompts to generate non-extractive characterizations of entities, as is evident from the examples. Examples of top characterizations of three test entities appearing across media houses are shown in Table ~\ref{tab:entitybymh} to contrast the characterizations generated by each media house. Generated characterizations have a cosine similarity score of greater than 0.75 with the FT1 corpus sentences. It is evident that top characterizations differ distinctly across media houses for the entities.

\section{Conclusion}
There are diverse perspectives about a person entity we know and even more with famous personalities. Media House discourses are diverse and impact the World Views of famous personalities. In today's world of the Information Age, getting insights into these World Views will lead to faster and better awareness. In this work, we propose an approach to derive common perceptions in a Zero-shot way. The evaluation criteria and metrics show a good performance of the approach.

\bibliographystyle{ACM-Reference-Format}
\bibliography{ech}


\begin{thebibliography}{21}


\ifx \showCODEN    \undefined \def \showCODEN     #1{\unskip}     \fi
\ifx \showDOI      \undefined \def \showDOI       #1{#1}\fi
\ifx \showISBNx    \undefined \def \showISBNx     #1{\unskip}     \fi
\ifx \showISBNxiii \undefined \def \showISBNxiii  #1{\unskip}     \fi
\ifx \showISSN     \undefined \def \showISSN      #1{\unskip}     \fi
\ifx \showLCCN     \undefined \def \showLCCN      #1{\unskip}     \fi
\ifx \shownote     \undefined \def \shownote      #1{#1}          \fi
\ifx \showarticletitle \undefined \def \showarticletitle #1{#1}   \fi
\ifx \showURL      \undefined \def \showURL       {\relax}        \fi
\providecommand\bibfield[2]{#2}
\providecommand\bibinfo[2]{#2}
\providecommand\natexlab[1]{#1}
\providecommand\showeprint[2][]{arXiv:#2}

\bibitem[Davison et~al\mbox{.}(2019)]%
        {davison-etal-2019-commonsense}
\bibfield{author}{\bibinfo{person}{Joe Davison}, \bibinfo{person}{Joshua
  Feldman}, {and} \bibinfo{person}{Alexander Rush}.}
  \bibinfo{year}{2019}\natexlab{}.
\newblock \showarticletitle{Commonsense Knowledge Mining from Pretrained
  Models}. In \bibinfo{booktitle}{\emph{Proceedings of the 2019 Conference on
  Empirical Methods in Natural Language Processing and the 9th International
  Joint Conference on Natural Language Processing (EMNLP-IJCNLP)}}.
  \bibinfo{publisher}{Association for Computational Linguistics},
  \bibinfo{address}{Hong Kong, China}, \bibinfo{pages}{1173--1178}.
\newblock
\urldef\tempurl%
\url{https://doi.org/10.18653/v1/D19-1109}
\showDOI{\tempurl}


\bibitem[Del~Corro and Gemulla(2013)]%
        {del2013clausie}
\bibfield{author}{\bibinfo{person}{Luciano Del~Corro} {and}
  \bibinfo{person}{Rainer Gemulla}.} \bibinfo{year}{2013}\natexlab{}.
\newblock \showarticletitle{Clausie: clause-based open information extraction}.
  In \bibinfo{booktitle}{\emph{Proceedings of the 22nd international conference
  on World Wide Web}}. \bibinfo{pages}{355--366}.
\newblock


\bibitem[Gao et~al\mbox{.}(2021)]%
        {gao-etal-2021-making}
\bibfield{author}{\bibinfo{person}{Tianyu Gao}, \bibinfo{person}{Adam Fisch},
  {and} \bibinfo{person}{Danqi Chen}.} \bibinfo{year}{2021}\natexlab{}.
\newblock \showarticletitle{Making Pre-trained Language Models Better Few-shot
  Learners}. In \bibinfo{booktitle}{\emph{Proceedings of the 59th Annual
  Meeting of the Association for Computational Linguistics and the 11th
  International Joint Conference on Natural Language Processing (Volume 1: Long
  Papers)}}. \bibinfo{publisher}{Association for Computational Linguistics},
  \bibinfo{address}{Online}, \bibinfo{pages}{3816--3830}.
\newblock
\urldef\tempurl%
\url{https://doi.org/10.18653/v1/2021.acl-long.295}
\showDOI{\tempurl}


\bibitem[Gururangan et~al\mbox{.}(2020)]%
        {gururangan2020don}
\bibfield{author}{\bibinfo{person}{Suchin Gururangan}, \bibinfo{person}{Ana
  Marasovi{\'c}}, \bibinfo{person}{Swabha Swayamdipta}, \bibinfo{person}{Kyle
  Lo}, \bibinfo{person}{Iz Beltagy}, \bibinfo{person}{Doug Downey}, {and}
  \bibinfo{person}{Noah~A Smith}.} \bibinfo{year}{2020}\natexlab{}.
\newblock \showarticletitle{Don’t Stop Pretraining: Adapt Language Models to
  Domains and Tasks}. In \bibinfo{booktitle}{\emph{Proceedings of the 58th
  Annual Meeting of the Association for Computational Linguistics}}.
  \bibinfo{pages}{8342--8360}.
\newblock


\bibitem[Hambardzumyan et~al\mbox{.}(2021)]%
        {hambardzumyan-etal-2021-warp}
\bibfield{author}{\bibinfo{person}{Karen Hambardzumyan}, \bibinfo{person}{Hrant
  Khachatrian}, {and} \bibinfo{person}{Jonathan May}.}
  \bibinfo{year}{2021}\natexlab{}.
\newblock \showarticletitle{{WARP}: {W}ord-level {A}dversarial
  {R}e{P}rogramming}. In \bibinfo{booktitle}{\emph{Proceedings of the 59th
  Annual Meeting of the Association for Computational Linguistics and the 11th
  International Joint Conference on Natural Language Processing (Volume 1: Long
  Papers)}}. \bibinfo{publisher}{Association for Computational Linguistics},
  \bibinfo{address}{Online}, \bibinfo{pages}{4921--4933}.
\newblock
\urldef\tempurl%
\url{https://doi.org/10.18653/v1/2021.acl-long.381}
\showDOI{\tempurl}


\bibitem[Jiang et~al\mbox{.}(2020)]%
        {jiang2020x}
\bibfield{author}{\bibinfo{person}{Zhengbao Jiang}, \bibinfo{person}{Antonios
  Anastasopoulos}, \bibinfo{person}{Jun Araki}, \bibinfo{person}{Haibo Ding},
  {and} \bibinfo{person}{Graham Neubig}.} \bibinfo{year}{2020}\natexlab{}.
\newblock \showarticletitle{X-FACTR: Multilingual Factual Knowledge Retrieval
  from Pretrained Language Models}. In \bibinfo{booktitle}{\emph{Proceedings of
  the 2020 Conference on Empirical Methods in Natural Language Processing
  (EMNLP)}}. \bibinfo{pages}{5943--5959}.
\newblock


\bibitem[Kassner and Sch{\"u}tze(2020)]%
        {kassner2020negated}
\bibfield{author}{\bibinfo{person}{Nora Kassner} {and} \bibinfo{person}{Hinrich
  Sch{\"u}tze}.} \bibinfo{year}{2020}\natexlab{}.
\newblock \showarticletitle{Negated and Misprimed Probes for Pretrained
  Language Models: Birds Can Talk, But Cannot Fly}. In
  \bibinfo{booktitle}{\emph{Proceedings of the 58th Annual Meeting of the
  Association for Computational Linguistics}}. \bibinfo{pages}{7811--7818}.
\newblock


\bibitem[Kumar and Talukdar(2021)]%
        {DBLP:conf/acl/KumarT21}
\bibfield{author}{\bibinfo{person}{Sawan Kumar} {and}
  \bibinfo{person}{Partha~P. Talukdar}.} \bibinfo{year}{2021}\natexlab{}.
\newblock \showarticletitle{Reordering Examples Helps during Priming-based
  Few-Shot Learning}. In \bibinfo{booktitle}{\emph{Findings of the Association
  for Computational Linguistics: {ACL/IJCNLP} 2021, Online Event, August 1-6,
  2021}} \emph{(\bibinfo{series}{Findings of {ACL}},
  Vol.~\bibinfo{volume}{{ACL/IJCNLP} 2021})},
  \bibfield{editor}{\bibinfo{person}{Chengqing Zong}, \bibinfo{person}{Fei
  Xia}, \bibinfo{person}{Wenjie Li}, {and} \bibinfo{person}{Roberto Navigli}}
  (Eds.). \bibinfo{publisher}{Association for Computational Linguistics},
  \bibinfo{pages}{4507--4518}.
\newblock
\urldef\tempurl%
\url{https://doi.org/10.18653/v1/2021.findings-acl.395}
\showDOI{\tempurl}


\bibitem[Lester et~al\mbox{.}(2021)]%
        {lester2021power}
\bibfield{author}{\bibinfo{person}{Brian Lester}, \bibinfo{person}{Rami
  Al-Rfou}, {and} \bibinfo{person}{Noah Constant}.}
  \bibinfo{year}{2021}\natexlab{}.
\newblock \showarticletitle{The power of scale for parameter-efficient prompt
  tuning}.
\newblock \bibinfo{journal}{\emph{arXiv preprint arXiv:2104.08691}}
  (\bibinfo{year}{2021}).
\newblock


\bibitem[Li and Liang(2021)]%
        {li2021prefix}
\bibfield{author}{\bibinfo{person}{Xiang~Lisa Li} {and} \bibinfo{person}{Percy
  Liang}.} \bibinfo{year}{2021}\natexlab{}.
\newblock \showarticletitle{Prefix-tuning: Optimizing continuous prompts for
  generation}.
\newblock \bibinfo{journal}{\emph{arXiv preprint arXiv:2101.00190}}
  (\bibinfo{year}{2021}).
\newblock


\bibitem[Liu et~al\mbox{.}(2021)]%
        {liu2021entity}
\bibfield{author}{\bibinfo{person}{Qingxia Liu}, \bibinfo{person}{Gong Cheng},
  \bibinfo{person}{Kalpa Gunaratna}, {and} \bibinfo{person}{Yuzhong Qu}.}
  \bibinfo{year}{2021}\natexlab{}.
\newblock \showarticletitle{Entity summarization: State of the art and future
  challenges}.
\newblock \bibinfo{journal}{\emph{Journal of Web Semantics}}
  \bibinfo{volume}{69} (\bibinfo{year}{2021}), \bibinfo{pages}{100647}.
\newblock


\bibitem[Liu et~al\mbox{.}(2020)]%
        {liu2020deeplens}
\bibfield{author}{\bibinfo{person}{Qingxia Liu}, \bibinfo{person}{Gong Cheng},
  {and} \bibinfo{person}{Yuzhong Qu}.} \bibinfo{year}{2020}\natexlab{}.
\newblock \showarticletitle{Deeplens: Deep learning for entity summarization}.
\newblock \bibinfo{journal}{\emph{arXiv preprint arXiv:2003.03736}}
  (\bibinfo{year}{2020}).
\newblock


\bibitem[Nishida et~al\mbox{.}(2020)]%
        {nishida2020unsupervised}
\bibfield{author}{\bibinfo{person}{Kosuke Nishida}, \bibinfo{person}{Kyosuke
  Nishida}, \bibinfo{person}{Itsumi Saito}, \bibinfo{person}{Hisako Asano},
  {and} \bibinfo{person}{Junji Tomita}.} \bibinfo{year}{2020}\natexlab{}.
\newblock \showarticletitle{Unsupervised Domain Adaptation of Language Models
  for Reading Comprehension}. In \bibinfo{booktitle}{\emph{Proceedings of the
  12th Language Resources and Evaluation Conference}}.
  \bibinfo{pages}{5392--5399}.
\newblock


\bibitem[Petroni et~al\mbox{.}(2019)]%
        {petroni2019language}
\bibfield{author}{\bibinfo{person}{Fabio Petroni}, \bibinfo{person}{Tim
  Rockt{\"a}schel}, \bibinfo{person}{Sebastian Riedel},
  \bibinfo{person}{Patrick Lewis}, \bibinfo{person}{Anton Bakhtin},
  \bibinfo{person}{Yuxiang Wu}, {and} \bibinfo{person}{Alexander Miller}.}
  \bibinfo{year}{2019}\natexlab{}.
\newblock \showarticletitle{Language Models as Knowledge Bases?}. In
  \bibinfo{booktitle}{\emph{Proceedings of the 2019 Conference on Empirical
  Methods in Natural Language Processing and the 9th International Joint
  Conference on Natural Language Processing (EMNLP-IJCNLP)}}.
  \bibinfo{pages}{2463--2473}.
\newblock


\bibitem[Radford et~al\mbox{.}(2019)]%
        {GPT2}
\bibfield{author}{\bibinfo{person}{Alec Radford}, \bibinfo{person}{Jeffrey Wu},
  \bibinfo{person}{Rewon Child}, \bibinfo{person}{David Luan},
  \bibinfo{person}{Dario Amodei}, \bibinfo{person}{Ilya Sutskever},
  {et~al\mbox{.}}} \bibinfo{year}{2019}\natexlab{}.
\newblock \showarticletitle{Language models are unsupervised multitask
  learners}.
\newblock \bibinfo{journal}{\emph{OpenAI blog}} \bibinfo{volume}{1},
  \bibinfo{number}{8} (\bibinfo{year}{2019}), \bibinfo{pages}{9}.
\newblock


\bibitem[Reynolds and McDonell(2021)]%
        {reynolds2021prompt}
\bibfield{author}{\bibinfo{person}{Laria Reynolds} {and} \bibinfo{person}{Kyle
  McDonell}.} \bibinfo{year}{2021}\natexlab{}.
\newblock \showarticletitle{Prompt programming for large language models:
  Beyond the few-shot paradigm}. In \bibinfo{booktitle}{\emph{Extended
  Abstracts of the 2021 CHI Conference on Human Factors in Computing Systems}}.
  \bibinfo{pages}{1--7}.
\newblock


\bibitem[Schick et~al\mbox{.}(2020)]%
        {schick2020automatically}
\bibfield{author}{\bibinfo{person}{Timo Schick}, \bibinfo{person}{Helmut
  Schmid}, {and} \bibinfo{person}{Hinrich Sch{\"u}tze}.}
  \bibinfo{year}{2020}\natexlab{}.
\newblock \showarticletitle{Automatically Identifying Words That Can Serve as
  Labels for Few-Shot Text Classification}. In
  \bibinfo{booktitle}{\emph{Proceedings of the 28th International Conference on
  Computational Linguistics}}. \bibinfo{pages}{5569--5578}.
\newblock


\bibitem[Schick and Sch{\"u}tze(2021a)]%
        {schick2021exploiting}
\bibfield{author}{\bibinfo{person}{Timo Schick} {and} \bibinfo{person}{Hinrich
  Sch{\"u}tze}.} \bibinfo{year}{2021}\natexlab{a}.
\newblock \showarticletitle{Exploiting Cloze-Questions for Few-Shot Text
  Classification and Natural Language Inference}. In
  \bibinfo{booktitle}{\emph{Proceedings of the 16th Conference of the European
  Chapter of the Association for Computational Linguistics: Main Volume}}.
  \bibinfo{pages}{255--269}.
\newblock


\bibitem[Schick and Sch{\"u}tze(2021b)]%
        {schick2021s}
\bibfield{author}{\bibinfo{person}{Timo Schick} {and} \bibinfo{person}{Hinrich
  Sch{\"u}tze}.} \bibinfo{year}{2021}\natexlab{b}.
\newblock \showarticletitle{It’s Not Just Size That Matters: Small Language
  Models Are Also Few-Shot Learners}. In \bibinfo{booktitle}{\emph{Proceedings
  of the 2021 Conference of the North American Chapter of the Association for
  Computational Linguistics: Human Language Technologies}}.
  \bibinfo{pages}{2339--2352}.
\newblock


\bibitem[Sukthanker et~al\mbox{.}(2020)]%
        {sukthanker2020anaphora}
\bibfield{author}{\bibinfo{person}{Rhea Sukthanker}, \bibinfo{person}{Soujanya
  Poria}, \bibinfo{person}{Erik Cambria}, {and} \bibinfo{person}{Ramkumar
  Thirunavukarasu}.} \bibinfo{year}{2020}\natexlab{}.
\newblock \showarticletitle{Anaphora and coreference resolution: A review}.
\newblock \bibinfo{journal}{\emph{Information Fusion}}  \bibinfo{volume}{59}
  (\bibinfo{year}{2020}), \bibinfo{pages}{139--162}.
\newblock


\bibitem[Wei et~al\mbox{.}(2019)]%
        {wei2019esa}
\bibfield{author}{\bibinfo{person}{Dongjun Wei}, \bibinfo{person}{Yaxin Liu},
  \bibinfo{person}{Fuqing Zhu}, \bibinfo{person}{Liangjun Zang},
  \bibinfo{person}{Wei Zhou}, \bibinfo{person}{Jizhong Han}, {and}
  \bibinfo{person}{Songlin Hu}.} \bibinfo{year}{2019}\natexlab{}.
\newblock \showarticletitle{ESA: entity summarization with attention}.
\newblock \bibinfo{journal}{\emph{arXiv preprint arXiv:1905.10625}}
  (\bibinfo{year}{2019}).
\newblock


\end{thebibliography}

\appendix

\end{document}